\documentclass{article}

\usepackage[preprint, nonatbib]{nips_2018}

\usepackage[utf8]{inputenc} 
\usepackage[T1]{fontenc}    
\usepackage{hyperref}       
\usepackage{url}            
\usepackage{booktabs}       
\usepackage{amsfonts}       
\usepackage{nicefrac}       
\usepackage{microtype}      
\usepackage{cite}

\usepackage{amsmath}
\usepackage{graphicx}
\usepackage{color}

\newcommand{\BiGraphNet}{$\texttt{BiGraphNet}$~}

\newcommand{\graph}[1]{\mathcal{#1}}
\newcommand{\set}[1]{\mathcal{#1}}
\newcommand{\RR}{I\!\!R} 
\newcommand{\bigO}{\mathcal{O}}
\DeclareMathOperator{\red}{\texttt{red}}
\DeclareMathOperator{\meanop}{\texttt{mean}}
\DeclareMathOperator{\sumop}{\texttt{sum}}
\DeclareMathOperator{\maxop}{\texttt{max}}
\DeclareMathOperator{\mlp}{\texttt{mlp}}
\DeclareMathOperator{\softmax}{\texttt{softmax}}

\title{Hierarchical Bipartite Graph Convolution Networks}

\author{
  Marcel ~Nassar \\
  Intel AI Lab\\
  Intel Artificial Intelligence Products Group\\
  \texttt{marcel.nassar@intel.com} \\
}

\begin{document}

\maketitle

\begin{abstract}
Recently, graph neural networks have been adopted in a wide variety of applications ranging from  relational representations to modeling irregular data domains such as point clouds and social graphs. However, the space of graph neural network architectures remains highly fragmented impeding the development of optimized implementations similar to what is available for convolutional neural networks. In this work, we present \BiGraphNet, a graph neural network architecture that generalizes many popular graph neural network models and enables new efficient operations similar to those supported by ConvNets.
By explicitly separating the input and output nodes, \BiGraphNet: 
(i) generalizes the graph convolution to support new efficient operations such as coarsened graph convolutions (similar to strided convolution in convnets), 
multiple input graphs convolution and graph expansions (unpooling) which can be used to implement various graph architectures such as graph autoencoders, and graph residual nets; and
(ii) accelerates and scales the computations and memory requirements in hierarchical networks by performing computations only at specified output nodes.
\end{abstract}
\section{Introduction}\label{sec:intro}
Convolutional neural networks (ConvNets) have been widely adopted in many applications from computer vision to NLP. At the core of ConvNets' success is the flexibility of the convolution operation (\textit{strided}, \textit{transposed}, and \textit{dilated}) and its adaptability to various seemingly different applications with different objectives such as localization vs invariance. 
Graph neural networks have been an active area of research \cite{Kipf2016,Bronstein2017GeometricGraphs,Battaglia2018RelationalNetworks}. While graph convnets share a lot with lattice convnets especially in terms of localized parameter sharing and the resulting stationarity, the repertoire of graph ConvNet operations remains limited to same-graph convolution followed by pooling. Despite having different flavors of graph convolutions, they don't typically modify the structure of the graph which is typically done by a separate pooling layer. In this work, we present a generalization of graph and lattice convolutions called \textit{bipartite} graph convolutions whose input and output vertex sets can be different, leading to new possible applications and scaling opportunities. 



\section{Graph Convolution Neural Networks}\label{sec:gconvnet}
In this section, we introduce the graph convolution operator and describe how some popular graph network architectures fit under this framework. In addition, we describe different graph coarsening and expansion operations that can be used to create hierarchical graph neural networks.

\subsection{Graph Convolution Operator}\label{sec:gconv-op}
A graph $\graph{G}$ is a tuple $(\set{V}, \set{E})$ denoted by $\graph{G}(\set{V}, \set{E})$ consisting of a vertex set $\set{V}=\{v_i\}_{i=1}^{N_\set{V}}$ and an edge set $\set{E}=\{e_j\}_{j=1}^{N_\set{E}}$. In weighted directed graphs, each edge $e_j$ is in turn a 3-tuple $(v, u, r)$ where $v$ is the source node, $u$ is the destination node, and $r$ is the edge label. On the other hand, each edge in undirected graphs can be seen as a 2-tuple $(\{v, u\}, r)$.

A graph \emph{signal} is a mapping $s: \set{V} \mapsto \RR^N $ such that $f_i=s(v_i)$ where $f_i$ is referred to as the node \emph{feature} of vertex $v_i$. A graph convolution operator $g: \graph{G}\times \RR^{|\set{V}|\times N} \mapsto \graph{G}\times \RR^{|\set{V}|\times M}$ uses the graph structure and locally  aggregates the graph signal as follows:
\begin{equation}\label{eq:gconv}
g_{\graph{G}}(v_i) = \red(\{W_{i,j}f_j| v_j \in \delta_{\graph{G}}(v_i), f_j=s(v_j)\})
\end{equation}
where $\red$ is a permutation-invariant reduction operation such as $\maxop$, $\meanop$, or $\sumop$. $\delta_{\graph{G}}(v_i)$ is the neighborhood of the node $v_i$ in $\graph{G}$. $W_{i,j} \in \RR^{M\times N}$ is a feature weighting kernel transforming the graph's $N$-dimensional features to $M$-dimensional ones. Fig.\ref{fig:graphconv}(a) illustrates the graph convolution operation being performed on node $v_1$ (in red): the features of the nodes in the $v_1$'s neighborhood $\delta(v_1)=\{v_2, v_3, v_4, v_5\}$ (in green) are multiplied by a kernel followed by a reduction operation.

\subsection{Weighting Kernels}\label{sec:wkernel}
Different network models differ in how they construct the weighting kernel $W_{i,j}$. Here, we discuss some popular examples (not meant to be exhaustive):
\paragraph{Edge Conditioned Kernels:}\label{par:ecc}
The weighting kernel $W_{i,j}$ is a parameterized function of the edge label between $v_i$ and  $v_j$ label (or relation) $r_{i, j}$; i.e. 
$W_{i, j} = k_{\theta}(r_{i,j})$
where $\theta$ are learnable parameters.
The parameterization of the kernel generation function varies: (i) in \cite{Simonovsky2017}, $k_{\theta}(\cdot)$ is chosen to be a neural network (typically an $\mlp$); (ii) in \cite{Monti2016}, on the other hand, $k_{\theta}(\cdot)$ is a mixture of Gaussians parameterized by their means and covariance matrices.  

\paragraph{Graph Attention Kernels:}\label{par:gat}
The graph attention kernel \cite{Velickovic2017} uses an attention mechanism \cite{Vaswani2017} to construct the weighting kernel as 
\begin{equation}
    W_{i,j} = \alpha_{i,j}W  \text{  where   } \alpha_{i,j}=\softmax(\mlp([Wf_i, W_fj])).
\end{equation}
In contrast to the edge conditioned kernel, this attention kernel uses node features for relational computations (concatenation of the two features), rather than the node’s relationship (if) given by the graph. 

\paragraph{Edge Conditioned Attention Kernel:} In order to make use of the graph's edge labels (relationships), we propose to combine the above two approaches to form an attention mechanism based on the edge (relationship) labeling $r_{i, j}$ as follows:
\begin{equation}
    W_{i,j} = \alpha_{i,j}W  \text{  where   } \alpha_{i,j}=\softmax(\mlp(r_{i, j})).
\end{equation}


\subsection{Hierarchical Graph Representations}\label{sec:hierarchy}
Hierarchical (also called \textit{multi-scale}) representations play a key role in the success of convolution neural networks. These representations can provide \emph{invariance} to local perturbations for classification tasks and encode information about larger structures that provide \emph{context} for the finer structures. 
Hierarchy is typically achieved through two operations: (i) pooling/downsampling and (ii) interpolation. ConvNets typically use a pooling kernel or strided convolution for downsampling, and a transposed convolution for interpolation, while graph neural nets require the additional steps of clustering and expanding the graph before such operations.


\subsubsection{Graph Pooling and Downsampling}\label{sec:pooling}
The pooling and downsampling procedure is illustrated in Fig.\ref{fig:graphconv}-(a,b,c). We discuss each step below.
\paragraph{Graph Clustering:}
constructs (or learns \cite{Ying2018}) a membership relationship mapping from each vertex $v_i\in \graph{G}$ into a set of groups $\set{C} = \{G_k\}_{k\in K}$ where $K \leq |V|$. Often, the clustering is vertex exclusive resulting in the new sub-graph having fewer nodes than the original. The clustering algorithm used depends on the type of graph data being processed: VoxelGrid \cite{Simonovsky2017} and Self Organizing networks \cite{Li2018} are typical for point cloud data, while Graclus \cite{Dhillon2007} can be used for general weighted graphs. It should be noted that dynamic pooling in the context of graphs is currently an active area of research \cite{Wang2018,Ying2018}.

\paragraph{Graph Pooling:} uses the graph clustering $\set{C}$ to group the "similar" nodes together into a new \emph{super-node} ($u_k$'s in Fig.\ref{fig:graphconv}-c). The feature of this super-node is computed as follows:
\begin{equation}\label{eq:pool}
    p_{\graph{G}}(u_k) = \red(\{f_j| v_j \in G_k, f_j=s(v_j)\})
\end{equation}
where $\red$ is the chosen reduction operation.

\subsubsection{Graph Expansion and Interpolation}\label{sec:expansion}
While \textit{interpolation} (upsampling using transpose convolution) is very common in many computer vision algorithms such as segmentation and super-resolution, it is not widely used in graph neural networks due to the ambiguity in performing graph expansions.  The most widely used application is the upsampling of point clouds such as in \cite{Fan2016,Yu2018PU-Net:Network}. These works focus on performing data-driven augmentation of the vertex sets representing the point cloud and don't make any explicit use of the graph structure. In \cite{Wang2018Pixel2Mesh}, the authors perform graph unpooling (form of graph expansion) by adding a vertex at the center of each edge and connecting it with the two end-point of this edge followed by a coordinate refinement step (interpolation).

\section{Bipartite Graph Convolution Neural Networks}\label{sec:bigraphnet}

\begin{figure}
  \centering
  \includegraphics[width=1.00\linewidth]{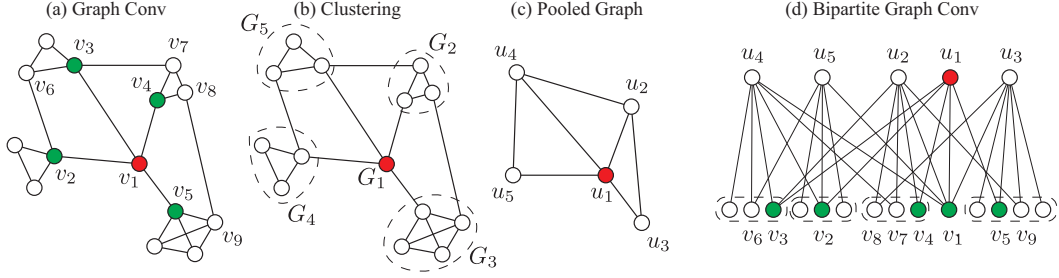}
  \caption{Typical graph convolution network: (a) apply graph conv operator on all nodes, (b) perform clustering, (c) followed by pooling vs. (d) a bipartite graph convolution directly performs a bipartite graph convolution from the input nodes to the output nodes}
  \label{fig:graphconv}
\end{figure}

A bipartite graph $\graph{BG}(\set{V},\set{U},\set{E})$ is a graph $\graph{G}(\set{V}\cup\set{U}, \set{E})$ where all the edges are between $\set{V}$ and $\set{U}$; i.e. $\set{E}=\{(v,u)|v\in\set{V}, u\in\set{U}\}$. 

\subsection{Bipartite Graph Convolution}\label{sec:rethink}
A bipartite graph convolution is defined over a bipartite graph $\graph{BG}(\set{V}_i, \set{V}_o, \set{E})$
\begin{equation}\label{eq:bigconv}
g_{\graph{BG}}(v_o) = \red(\{W_{o,i}f_i| v_i \in \delta_{\graph{BG}}(v_o), f_i=s(v_i)\}) \text{    } \forall v_o \in \set{V}_o
\end{equation}
where $\red$ is a reduction operation and  $\delta_{\graph{GB}}(v_o)=\{v_i\in\set{V}_i|(v_i, v_o)\in\set{E}\}$ is the neighborhood of the node $v_o$ in $\graph{BG}$. $W_{o,i} \in \RR^{M\times N}$ is a feature weighting kernel. Equation (\ref{eq:bigconv}) shows that the domain of the bipartite graph convolution is given by the set $\set{V}_i$, while its co-domain is given by the set $\set{V}_o$ (see Fig.\ref{fig:graphconv}(d)). As a result, the output nodes are directly connected to the input nodes and the computation is executed only on the desired output nodes rather than on all the nodes of the input graph. Given that any graph $\graph{G}(\set{V}, \set{E})$ can be written as a bipartite graph $\graph{BG}(\set{V}, \set{V}, \set{E})$, it is trivial to write a graph convolution on $\graph{G}$ as a bipartite graph convolution on the corresponding $\graph{BG}$.

\subsection{Hierarchical Bipartite Convolutions}
Consider two sequential layers: a graph convolution layer operating on a graph $\graph{G}_i(\set{V}_i, \set{E}_i)$  followed by a graph pooling layer resulting in a output graph $\graph{G}_o(\set{V}_o, \set{E}_o)$. Using a bipartite graph convolution, a similar mapping can be learned on $\graph{BG}(\set{V}_i, \set{V}_o, \set{E}^{'})$ where $\set{E}^{'}$ includes all the edges corresponding to $\set{E}_o$ and the edges in $\set{E}_i$ remapped to the super-nodes $u_k$ (see Fig.\ref{fig:graphconv}-d). By specifying the output node set, we can implement graph downsampling and interpolation as described in sections \ref{sec:pooling} and \ref{sec:expansion}. Note that, in general, graph conv followed by pooling is not equivalent to a bipartite graph convolution due to the presence of the $\maxop$; however, it implements operations that are similar to strided, dilated, and transpose convolution in ConvNets. In fact, when treating the graphs as regular lattices such as in images, these three types of convolutions are a special case of the bipartite graph convolution.

\paragraph{Computational Complexity and Memory Requirements:} Since that the computations are executed only on the output nodes, the computational complexity of the forward pass now scales as $\bigO(|\set{V}_o|)$ instead of the $\bigO(|\set{V}_i|+|\set{V}_o|)$ required to execute same graph convolution followed by graph pooling (see Fig.\ref{fig:graphconv}(a-c)). Memory requirements follow a similar trend. In typical applications such as point cloud classification where $|\set{V}_o|\ll|\set{V}_i|$ this results in huge computational and memory savings.

\begin{figure}
  \centering
  \includegraphics[width=0.850\linewidth]{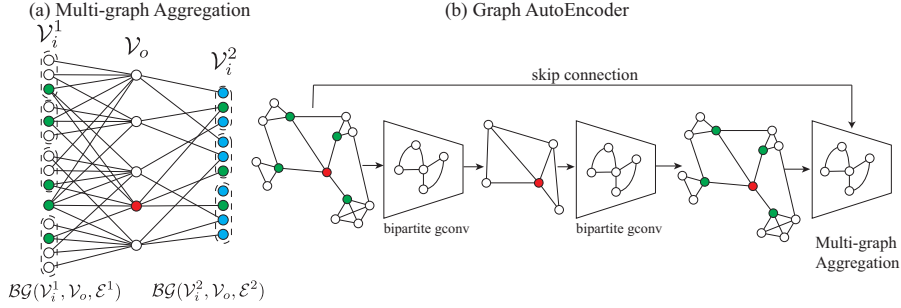}
  \caption{(a) Aggregation of two graphs with vertex sets $\set{V}_i^1$ and $\set{V}_i^2$  into the same output $\set{V}_o$, (b) part of a graph autoencoder where a multi-graph aggregation layer is used to implement a skip connection}
  \label{fig:autoencoder}
\end{figure}

\subsection{Multiple Graph Aggregation and Graph AutoEncoders}
By specifying the same output vertex set $\set{V}_o$, the bipartite graph convolution can be used to aggregate inputs from different input graphs $\graph{BG}_1(\set{V}_i^1, \set{V}_o, \set{E}^1)$ and $\graph{BG}_2(\set{V}_i^2, \set{V}_o, \set{E}^2)$ (see Fig.\ref{fig:autoencoder}(a)) as
\begin{equation}
    g_{aggr}(v_o) = \frac{1}{|\set{V}_i^1|+ |\set{V}_i^2|} [|\set{V}_i^1|\times g_{\graph{BG}_1}(v_o) +  |\set{V}_i^2|\times g_{\graph{BG}_2}(v_o)]
\end{equation}
This can be used to fuse multiple relationship graphs or to implement graph skip connections as shown in Fig.\ref{fig:autoencoder}. 

\section{Experiments}

\paragraph{Architecture Specification:} we denote a graph convolution layer as C($x$) where $x$ denotes the feature dim, a max-pooling layer as MP($r$, $\rho$) where $r$ is radius connectivity and $\rho$ is the resolution, a fully connected layers as FC($y$) where $y$ is the number of output neurons, and a global max-pooling as GMP.

\paragraph{Point Cloud Classification:}\label{sec:pcl-class}
We test the performance of the \BiGraphNet architecture on the ModelNet10 benchmark \cite{Wu2015ShapeNets}. ModelNet10 is a large-scale collection of mesh surfaces of $10$ object categories such as tables and chairs split into $3991$ train examples and $908$ test examples. Each of the example meshes is sampled uniformly to produce $1000$ points. In our experiments, we use the dataset available at \url{https://github.com/mys007/ecc} \cite{Simonovsky2017}. Each example point cloud is centered and normalized such that it is within the unit sphere (i.e., each coordinate $x, y, z\in[-1, 1]$). No further augmentations were used. We use the classification architecture given in \cite{Simonovsky2017}: C(16)-C(32)-MP(2.5/32,7.5/32)-C(32)-C(32)-MP(7.5/32,22.5/32)-C(64)-GMP-FC(64)- D(0.2)-FC(10). The \BiGraphNet architecture uses the same spatial parameters but eliminates the pooling layers as they are not needed in bipartite graph convolutions.
The results are given in Table~\ref{tab:modelnet}. The \BiGraphNet architecture achieves better performance than the corresponding ECC architecture. 

\begin{table}\label{tab:modelnet}
\begin{center}
\begin{tabular}{|l|c|c|}
\hline
Model & Precision \\
\hline\hline
VoxNet& $92.0$  \\
ORION& $93.8$  \\
RotationNet & $98.46$ \\
ECC & $89.3$ \\
\hline
BGN-FullConv & $92.95$\\
\hline
\end{tabular}
\end{center}
\caption{ModelNet10 classification results}
\end{table}

\paragraph{MNIST as a graph:} 
Each image ($\set{I}$) can be interpreted as a signal (given by the pixel intensity) defined over a set of coordinate nodes $\set{P}=\{(x,y)|x, y\in \{0, \cdots,27\}\}$. We follow the setup of \cite{Simonovsky2017} and use the spatial neighborhood to define a relationship between two (abstract) nodes $v_i, v_j$ as follows
\begin{math}
        v_i \xrightarrow[]{r_{ij}} v_j \text{ with } r_{ij}=p_i-p_j \text{ iff } p_j \in \delta_{\rho}(p_i).
\end{math}
where $\delta_\rho(\cdot)$ represents the spatial neighborhood of radius $\rho=2.9$.
Graph coarsening is implemented using the VoxelGrid algorithm \cite{Simonovsky2017}.

\paragraph{Classification}
We test the proposed bigraphnet against the ECC MNIST network given in \cite{Simonovsky2017}. The network has the following architecture: C(16)-MP(2,3.4)-C(32)-MP(4,6.8)-C(64)-MP(8,30)-C(128)-D(0.5)-FC(10). We tested both max and avg pooling and compared to the  \BiGraphNet architecture without the pooling layers. The results after $20$ epochs: ECC-avg@20: $97.59$, \BiGraphNet @20: $97.71$, ECC max@20: $99.05$. The \BiGraphNet outperforms the avg pooling ECC but still doesn't match the max pooling which adds invariance for classification.

\begin{table}[h!]
  \caption{Graph Autoencoder Test MSE Results}
  \label{tab:sample-table}
  \centering
  \begin{tabular}{llllllll}
    \toprule
    \cmidrule(r){1-2}
    \midrule
    MLP & $0.088$ &  ConvNet & $0.094$ & Graph & $0.086$ & Skipped Graph& $0.066$\\
    \bottomrule
  \end{tabular}
\end{table}

\paragraph{Graph Autoencoder}
We also test graph autoencoder that makes use the bigraphnet to upsample the graph to a pixel grid and the multiple graph aggregations to implement skip connections. We compare against an MLP and ConvNet Autoencoders. The results are given in Table\ref{tab:sample-table}.



\bibliographystyle{plain}
\bibliography{refs}

\begin{thebibliography}{10}

\bibitem{Battaglia2018RelationalNetworks}
Peter~W. Battaglia and et. al.
\newblock {Relational inductive biases, deep learning, and graph networks}.
\newblock 2018.

\bibitem{Bronstein2017GeometricGraphs}
Michael Bronstein, Xavier Bresson, Yann Lecun, Arthur Szlam, and Joan Bruna.
\newblock {Geometric deep learning on graphs}.
\newblock {\em IEEE Signal Processing Magazine}, (July), 2017.

\bibitem{Dhillon2007}
Inderjit~S. Dhillon, Yuqiang Guan, and Brian Kulis.
\newblock {Weighted Graph Cuts without Eigenvectors: A Multilevel Approach}.
\newblock {\em PAMI}, 2007.

\bibitem{Fan2016}
Haoqiang Fan, Hao Su, and Leonidas Guibas.
\newblock {A Point Set Generation Network for 3D Object Reconstruction from a
  Single Image}.
\newblock {\em CVPR}, 2017.

\bibitem{Kipf2016}
Thomas~N. Kipf and Max Welling.
\newblock {Semi-Supervised Classification with Graph Convolutional Networks}.
\newblock {\em ICLR}, 2017.

\bibitem{Li2018}
Jiaxin Li, Ben~M Chen, and Gim~Hee Lee.
\newblock {SO-Net: Self-Organizing Network for Point Cloud Analysis}.
\newblock {\em CVPR}, 2018.

\bibitem{Monti2016}
Federico Monti, Davide Boscaini, Jonathan Masci, Emanuele Rodol{\`{a}}, Jan
  Svoboda, and Michael~M Bronstein.
\newblock {Geometric deep learning on graphs and manifolds using mixture model
  CNNs}.
\newblock {\em CVPR}, 2016.

\bibitem{Simonovsky2017}
Martin Simonovsky and Nikos Komodakis.
\newblock {Dynamic Edge-Conditioned Filters in Convolutional Neural Networks on
  Graphs}.
\newblock {\em CVPR}, 2017.

\bibitem{Vaswani2017}
Ashish Vaswani, Noam Shazeer, Niki Parmar, Jakob Uszkoreit, Llion Jones,
  Aidan~N. Gomez, Lukasz Kaiser, and Illia Polosukhin.
\newblock {Attention Is All You Need}.
\newblock {\em NIPS}, 2017.

\bibitem{Velickovic2017}
Petar Veli{\v{c}}kovi{\'{c}}, Guillem Cucurull, Arantxa Casanova, Adriana
  Romero, Pietro Li{\`{o}}, and Yoshua Bengio.
\newblock {Graph Attention Networks}.
\newblock {\em ICLR}, 2018.

\bibitem{Wang2018Pixel2Mesh}
Nanyang Wang, Yinda Zhang, Zhuwen Li, Yanwei Fu, Wei Liu, and Yu-Gang Jiang.
\newblock {Pixel2Mesh: Generating 3D Mesh Models from Single RGB Images}.
\newblock 2018.

\bibitem{Wang2018}
Yue Wang, Yongbin Sun, Ziwei Liu, Sanjay~E. Sarma, Michael~M. Bronstein, and
  Justin~M. Solomon.
\newblock {Dynamic Graph CNN for Learning on Point Clouds}.
\newblock 2018.

\bibitem{Ying2018}
Rex Ying, Jiaxuan You, Christopher Morris, Xiang Ren, William~L Hamilton, and
  Jure Leskovec.
\newblock {Hierarchical Graph Representation Learning with Differentiable
  Pooling}.
\newblock {\em NIPS}, 2018.

\bibitem{Yu2018PU-Net:Network}
Lequan Yu, Xianzhi Li, Chi-Wing Fu, Daniel Cohen-Or, and Pheng-Ann Heng.
\newblock {PU-Net: Point Cloud Upsampling Network}.
\newblock {\em CVPR}, 2018.

\bibitem{Wu2015ShapeNets}
A.~Khosla F. Yu L. Zhang X.~Tang Z.~Wu, S.~Song and J.~Xiao.
\newblock {3D ShapeNets: A Deep Representation for Volumetric Shape Modeling}.
\newblock {\em CVPR}, 2015.

\end{thebibliography}

\end{document}